\documentclass[10pt,twocolumn,letterpaper]{article}

\usepackage{cvpr}
\usepackage{times}
\usepackage{epsfig}
\usepackage{graphicx}
\usepackage{amsmath}
\usepackage{amssymb}
\usepackage{multirow}
\usepackage{array}

\usepackage[pagebackref=true,breaklinks=true,letterpaper=true,colorlinks,bookmarks=false]{hyperref}

\cvprfinalcopy 

\def\cvprPaperID{13} 

\ifcvprfinal\fi
\begin{document}

\title{Effect of Annotation Errors on Drone Detection with YOLOv3
}

\author{Aybora Koksal, Kutalmis Gokalp Ince, A. Aydin Alatan\\
Department of Electrical and Electronics Engineering, Center for Image Analysis (OGAM)\\
Middle East Technical University, Ankara, Turkey\\
{\tt\small aybora, kutalmis, alatan @ metu.edu.tr}
}

\maketitle

\begin{abstract}
   Following the recent advances in deep networks, object detection and tracking algorithms with deep learning backbones have been improved significantly; however, this rapid development resulted in the necessity of large amounts of annotated labels. Even if the details of such semi-automatic annotation processes for most of these datasets are not known precisely,  especially for the video annotations, some automated labeling processes are usually employed. Unfortunately, such approaches might result with erroneous annotations. In this work, different types of annotation errors for object detection problem are simulated and the performance of a popular state-of-the-art object detector, YOLOv3, with erroneous annotations during training and testing stages is examined. Moreover, some inevitable annotation errors in Anti-UAV Challenge dataset is also examined in this manner, while proposing a solution to correct such annotation errors of this valuable data set.
\end{abstract}


\vspace{-1em}
\section{Introduction}
Object detection literature has been developed very rapidly throughout the last couple of years. After Convolutional Neural Networks (CNNs) become popular, conventional feature extraction methods are replaced with convolutional layers. Based on deep CNN structures, various object detectors are proposed~\cite{Girshick1,Girshick2,Ren,He}, including one-shot detectors~\cite{yolo,ssd,yolo9000,retinanet}.

A typical one-shot detector generally finds all the defined class objects with their bounding box information and objectness scores. After this step, some post processing stage might be employed, such as non-maxima suppression, to eliminate duplicated results. Since these methods are designed to perform on each image independently, they can be employed for tracking problems as well without any drift problem. After recent developments in GPU technology and efficiency enhancements of one-shot detectors, there are alternative methods working in near real-time ~\cite{yolo9000,retinanet} that made them to be employed them in real-time tracking problems.

YOLOv3~\cite{yolov3} is a popular state-of-the-art object detector belonging to one-shot object detector family that works close to real-time and more accurate compared to the previous approaches ~\cite{ssd,retinanet}. Moreover, lots of Tensorflow and PyTorch implemented versions of YOLOv3 are available in addition to its original version, which is implemented in C, with an API called Darknet. 

Similar to the other CNN-based object detectors, YOLOv3 also requires a large amount of labeled data for the training process which requires  significant amount of manpower. Especially for video annotations, labeling a small number of images by hand and interpolating the intermediate frames with tracking might be an acceptable idea to save manpower and time. Unfortunately, since these algorithms are not perfect, there might be a discrepancy between the real data and interpolated data which results in annotation errors.

In this study, YOLOv3 algorithm is firstly trained with CVPR-2020 Anti-UAV Challenge dataset by fine tuning the existing weights of the algorithm to detect drone classes. YOLOv3 is trained for different number of drone classes and different number of epochs with different amount of data to figure out the most efficient way of training in terms of training time and performance. Next, the tracking accuracy of YOLOv3 technique is analyzed by considering the provided annotations.

After selection of the best way for training, some additional annotation errors are applied to the dataset in order to create separate new datasets each consisting different type of errors and some combined ones. Then, YOLOv3 is trained with each of these new erroneous datasets and the results are compared in terms of precision, recall and tracking accuracy.

Since some incorrect annotations are observed in CVPR-2020 Anti-UAV Challenge dataset, a novel semi-automatic approach is also proposed to correct erroneous annotations to improve the labeling accuracy of this valuable dataset. Moreover, the accuracy between corrected and original labels are calculated in terms of mean and standard deviation.


This paper has three main objectives: The first objective is to reveal the performance of a state-of-the-art detector for small objects which can serve as a baseline for detection-based tracking methods. The second objective is to investigate the performance of the detector in the presence of annotation errors. The final objective of this paper is to come up with a semi-automatic method to correct such annotation errors that are already present in CVPR-2020 Anti-UAV Challenge dataset.

The rest of the paper is organized as follows: Firstly, related work on deep learning-based object detectors and training with noisy data are presented. Section 3 and 4 are dedicated to training of YOLOv3 with Anti-UAV Challenge dataset and semi-automatic annotation correction. In Experiments section, the results for original, noisy and corrected datasets are compared. Conclusions of the experimental evidence are presented in the last section.

\section{Related Work}

\subsection{Deep Learning-based Object Detection}

In the literature, deep learning-based object detectors are mostly classified  into two classes: Two-stage (region proposal based) and one-stage detectors ~\cite{Jiao,Liu}. On the other hand, there are few-shot learning algorithms which are mostly used for object counting or segmentation ~\cite{wang2019generalizing,zhao2018dynamic}.

\begin{figure}[b]
\centering
\vspace{-1em}
   \includegraphics[width=1\linewidth]{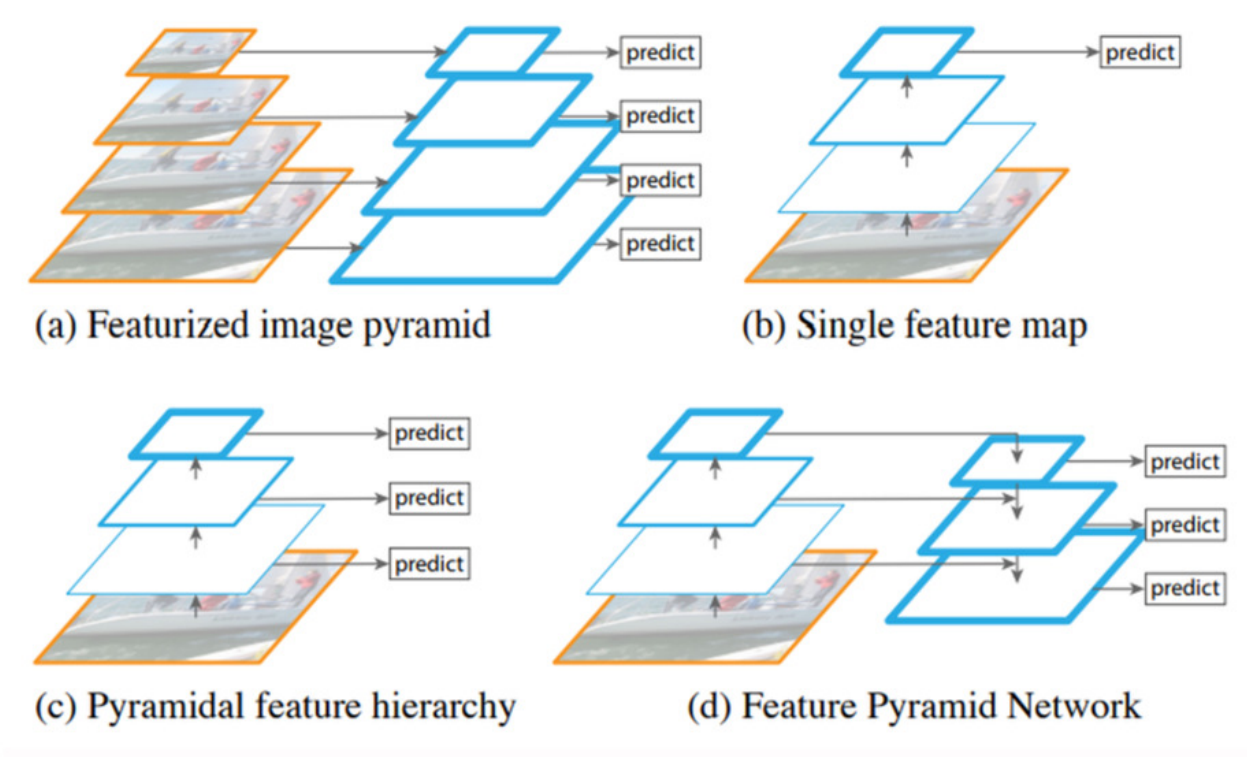}
   \caption{Feature pyramid alternatives. (a) Most straight forward solution since every size has its own network. Therefore, the network is very slow. Used in~\cite{adelson1984pyramid}. (b) One prediction at the end of the network. Gradients might vanish for small objects. Used in ~\cite{hespatial2014,Girshick2,Ren}. (c) Different predictions for different layers but previous layer prediction cannot use deeper layer information. Used in~\cite{ssd} (d) Feature pyramid network. Taken from ~\cite{fpn} where this method is also proposed.}
\label{fig:fpn}
\end{figure}

R-CNN~\cite{Girshick1} can be considered as the first two-stage detector proposed for object detection. This algorithm shows that a deep learning-based object detector can rapidly increase the performance in PASCAL-VOC dataset~\cite{everingham2009}. Girshick et al.~\cite{Girshick2} then proposed a faster version of R-CNN, namely Fast R-CNN. Instead of extracting features from each region proposal separately, Fast R-CNN perform the feature extraction on entire image and propagate them to the region of interest (RoI) pooling layer. Ren et al.~\cite{Ren} later  proposed Faster R-CNN, which improves region proposal architecture. Fast R-CNN uses selective search to find related region proposals, which decreases detection time dramatically. Meanwhile, Faster R-CNN introduces a completely new structure, denoted as Region Proposal Network (RPN) for generating proposals. RPN is a fully convolutional structure which predicts region proposals in different scales or sizes, whereas Fast R-CNN consists fully connected layers which weakens the network considerably. The idea behind feature mapping on RPN is visually presented in Figure~\ref{fig:fpn} (b). He et al.~\cite{He} introduced Mask R-CNN as an extension of Faster R-CNN to create a framework for object instance segmentation. In Mask R-CNN method, ResNet-FPN~\cite{fpn} (feature pyramid network) is utilized for the feature extraction backbone and such an approach helps the algorithm to achieve higher accuracy with reduced computation time. FPN structure is presented in Figure~\ref{fig:fpn} (d).


As a pioneering one-stage object detector, Redmon et al.~\cite{yolo} proposed YOLO (You Only Look Once)algorithm which works as a real-time application. The reason for its real time performance is due to the fact that instead of doing selective search for thousands of region proposals, as in Fast R-CNN, YOLO simply predicts less than 100 bounding boxes for each image. Another one-stage object detector, SSD (single shot detector)~\cite{ssd} is proposed by Liu et al., works as a multiple class object detector, which determines class scores and bounding boxes from a fixed set of bounding boxes of different sizes and scales. SSD combines ideas from RPN of Faster R-CNN and YOLO; moreover, it also adds multiscale convolutional layers for feature extraction to increase detection speed while preserving accuracy. 

Redmon et al.~\cite{yolo9000} later improved their work YOLO with a newer version, YOLOv2 which utilizes a completely new feature extractor backbone, called Darknet19, since it consists 19 convolutional layers. In YOLOv2, fully connected layers are removed and convolutional layers are used to predict bounding boxes. In 2018, RetinaNet~\cite{retinanet} is proposed by Lin et al. as another one-stage object detector. The main novelty of RetinaNet algorithm is reducing the class imbalance effect between foreground and background of each object which causes two-stage object detectors having higher precision than the one-stage ones. In order to gain robustness against class imbalance, Lin et al. proposed a new loss function, namely focal loss, which reduces the weight of easy examples during training. 

Redmon et al.~\cite{yolov3} further improved their algorithm into a new version, namely YOLOv3. YOLOv3 enables multi-class detection by using logistic loss function instead of softmax layer, since there could be possible cases for which an object belong to more than one class. Based on feature pyramid network, which is illustrated in Figure~\ref{fig:fpn} (d), YOLOv3 gives three different level of detection for three various sized objects.

In order to compete with trackers, in this study only one shot detectors are considered that work in real-time or near-real-time. According to the results in~\cite{Jiao}, on MS COCO dataset~\cite{coco} YOLOv3 achieves 57.9\% mAP, meanwhile RetinaNet has 61.1\% mAP, meanwhile YOLOv3 operates nearly 4 times faster than RetinaNet. Moreover, YOLOv3 is a better alternative for small objects (hence with drones), since it uses multi-scale detection. It provides nearly real time object detector with good performance potential on small objects, YOLOv3 is selected for the erroneous annotation experiments.

\subsection{Training with Erroneous Annotations for Object Detection}

Labeling errors about the training data is already examined in the object detection literature. Frenay et al.~\cite{frenay} defined annotation errors as an independent stochastic process which may or may not be introduced intentionally. The authors have done a detailed survey that includes learning in presence of labeling noises, such as some probabilistic models which are Bayes-optimal classifiers~\cite{perez}. Moreover, they included some semi/weakly supervised methods~\cite{breve} that prevent mislabelled instances from affecting detection performance considerably. Moreover, they examined some noise-cleansing algorithms, such as detection of mislabelled instances by using class confidence metrics~\cite{sun}.

Rolnick et al.~\cite{rolnick2017deep} argues that introducing label noise into a training set reduces the performance of CNNs, although it is not as remarkable as the multi-layer perceptron networks. In addition to this argument, the authors also stated that more deeper networks, such as ResNet~\cite{resnet}, are less affected from such a noise. Moreover, the authors conclude that to attain the same accuracy level, the training set with higher rate of noisy labels need to be larger.

Noisy labels can also be a problem for weakly supervised object segmentation tasks. Lu et al.~\cite{zlu} introduces a superpixel noise reduction algorithm which is based on a sparse learning model. Next, with this cleaned labels, an iterative superpixel label prediction/appearance model is created. Using this method, the authors increased total per-pixel accuracy by 5 to 15\% in comparison to the best other method~\cite{xuweak}.

\begin{figure}[h!]
\centering
   \includegraphics[width=1\linewidth]{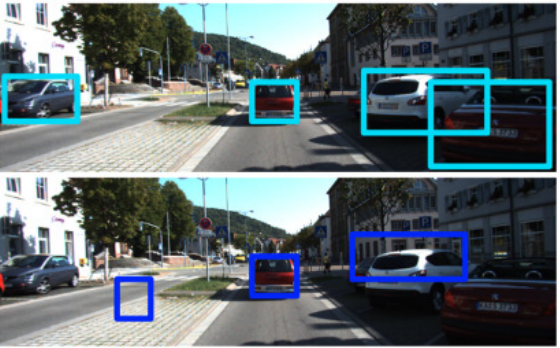}
   \caption{Simple example of noisy labeling on KITTI dataset. First image shows real ground truth labeling while the latter one shows some noises~\cite{Chadwick}.}
\label{fig:noisy_kitti}
\end{figure}

In one of the most related and recent research ~\cite{Chadwick}, the authors trained their SSD-based framework with KITTI dataset~\cite{kitti} and artificial annotation errors which are additional boxes, missing boxes and shifted boxes. A typical visual of a sample annotation error on KITTI dataset is presented in Figure~\ref{fig:noisy_kitti}. The performance of SSD with or without annotation errors are also reported as shown in Table~\ref{tab:kitti}. According to the results, additional boxes decreases performance, that decrease are not related with the noise probability. Missing and shifted boxes on the other hand, decreases precision further with increasing noise probability, with similar rates. Upon all of the noise types, combined labeling noise affects the network most, as expected.

\vspace{-1em}

\begin{table}[h!]
\centering
\caption{Performance on KITTI dataset in terms of average precision with different types of noises with varying levels.~\cite{Chadwick}.}
\label{tab:kitti}
\medskip
\begin{tabular}{|l|c|c|c|}
\hline
& \multicolumn{3}{|c|}{Noise Probability} \\
\hline
Noise Type & 0.0 & 0.25 & 0.5 \\
\hline
No Noise & 0.629 & - & - \\
Additional boxes & - & 0.560 & 0.587 \\
Missing Boxes & - & 0.593 & 0.518\\
Shifted Boxes & - & 0.577 & 0.502\\
Combined & - & 0.457 & 0.317\\
\hline
\end{tabular}
\end{table}

\vspace{-1em}


\section{Performance Metrics}
Since Anti-UAV Challenge dataset is aimed for the tracking problem, the performance metric for this challenge is announced as the average intersection over union based on the assumption that there is at most one output object on each frame. However, in this paper a detection algorithm is studied; therefore, additional performance metrics are required. Hence, for this purpose \textit{hit rate} and \textit{number of false alarms} are also evaluated as the additional performance metrics. If a detection output has IoU larger than 0.5 for the annotated object, then this result is counted as a hit (Pascal criteria). In case of zero IoU, the decision is counted as false alarm. Finally, for non-zero IoU smaller than 0.5, no additional penalty is applied as the annotated object is missed and penalty is already included in the hit rate. No detection output for no annotation, i.e. true rejection, is not counted. For rest of the paper, comparison results are presented in terms of false alarms per minute and hit rate, in addition to the tracking accuracy metric given in Anti-UAV Challenge, $TA$ which is defined as:  
\begin{equation}
TA = \frac{1}{T} \sum_{t=1}^{T} IoU_t*v_t*p_t+(1-p_t)(1-v_t) \label{eq:1}
\end{equation}
where $T$ is number frames, $IoU_{t}$ is intersection over union, $v_{t} \in \{0, 1\}$ is visibility flag, and $p_{t} \in \{0, 1\}$ is prediction flag at frame $t$.

Since the object detector YOLOv3 might generate more than one detection result on a single frame, tracking accuracy metric in Eq.~\ref{eq:1} cannot penalize additional false alarms. Therefore, this metric is also slightly modified so that the false alarms reduce the accuracy. We define this modified tracking accuracy as follows to penalize additional false alarms:

\begin{equation}
MTA = \frac{\sum_{t=1}^{T} {IoU_t*v_t*p_t+(1-p_t)*(1-v_t)}}
{\sum_{t=1}^{T} {max(v_t, p_t) + (1-p_t)*(1 - v_t)}}
\label{eq:2}
\end{equation}
This modified tracking accuracy is equal to original tracking accuracy as long as the number of detection per frame is limited to one, but each additional false detection reduces the tracking accuracy. 

\section{Training YOLOv3 with Anti-UAV Dataset}

\textbf{Modified Network:} YOLOv3 network is pretrained to detect 80 different classes, while the input image is divided into grids on three different scales. For each grid cell in each scale, YOLOv3 generates a vector containing the objectness score, class probabilities and bounding box for three alternative anchor boxes. Therefore, for each cell the length of the output vector is 3x(1+80+4)=255. For drone detection, we have trained YOLOv3 only for one- and three-class alternatives resulting in output vectors of length 3x(1+1+4)=18 and 3x(1+3+4)=24, respectively. For the one-class case, the network is trained only with thermal images to detect drones. For three-class case, the network is trained with RGB day, RGB night and thermal images which correspond to three different drone classes. The performance of these two alternatives are compared to understand whether there is a significant difference between one class and three class cases or not.

\textbf{Dataset:} For the thermal image dataset, "test-dev" part of Anti-UAV Challenge dataset is used. RGB videos are also included for three-class scenario to examine whether the including them increases the accuracy or not. The videos are divided randomly as training and validation set with weights of 70\% and 30\%, respectively. 

\textbf{Training:} During the training, different dataset sizes and different epoch numbers tested for one- and three-class alternatives. Since for each annotation error, simulation of the network should be trained again, precision/training time efficiency is considered for comparison. The results of 25, 50, 100th epochs with full dataset, half dataset which is obtained by getting one frame and skipping the next one, and one quarter dataset which is obtained by getting one frame and skipping the next three, are also compared.

As tabulated in Table~\ref{tab:thermalvsrgb}, the trained network produces quite similar results for one-class and three-class cases for the full dataset, whereas training time is extended twice for the three-class scenario. Hence for the rest of the paper, we only focused on one-class case. Moreover, at 100th epoch, false alarms are increased due to some memorizing or overfitting. Therefore, for the rest of the paper, only 25th and 50th epochs are compared. As presented in Table~\ref{tab:fullvshalf}, the most efficient performance is on 50th epoch for half dataset. Since data from the adjacent video frames are quite redundant, removing half of the dataset does not decrease performance of the network. On the other hand, using only quarter of the dataset decreases the performance. However, it is difficult to deduce whether this result is due to either losing data variety or number of samples in the set. To sum up, for the rest of the paper, the network is trained for one-class only with the half of the thermal images for 50 epochs. 

\begin{table}[h!]
\centering
\caption{Performance comparison of YOLOv3 on Thermal Test Set when trained only with thermal data (one-class) vs Thermal+RGB data (three-classes) in terms of Hit Rate (\%) and False Alarm (per minute)}
\label{tab:thermalvsrgb}
\medskip
\begin{tabular}{|l|c|c|c|c|c|c|}
\hline
\# Epoch & \multicolumn{2}{|c|}{25} & \multicolumn{2}{|c|}{50} & \multicolumn{2}{|c|}{100} \\
\hline
 & HR & FA & HR & FA & HR & FA \\
\hline
Thermal & 97.5 & 2.4 & 97.1 & 2.2 & 97.3 & 3.5 \\
Thermal+RGB & 96.9 & 2.3 & 97.4 & 1.7 & 97.9 & 4.3 \\
\hline
\end{tabular}
\end{table}

\vspace{-1em}

\begin{table}[h]
\centering
\caption{Performance comparison of YOLOv3 on Thermal Test Set for different number of epochs and different dataset sizes in terms of Hit Rate (\%) and False Alarm (per minute)}
\label{tab:fullvshalf}
\medskip
\begin{tabular}{|l|c|c|c|c|}
\hline
\# Epoch & \multicolumn{2}{|c|}{25} & \multicolumn{2}{|c|}{50} \\
\hline
 & HR & FA & HR & FA \\
\hline
Full dataset & 97.5 & 2.4 & 97.1 & 2.2 \\
1/2 dataset & 95.7 & 2.4 & 97.5 & 2.4 \\
1/4 dataset & 93.9 & 2.7 & 95.1 & 2.1 \\
\hline
\end{tabular}
\end{table}

\vspace{-1em}


\section{Annotation Errors in Anti-UAV Dataset}

\begin{figure}[t]
\centering
   \includegraphics[width=1\linewidth]{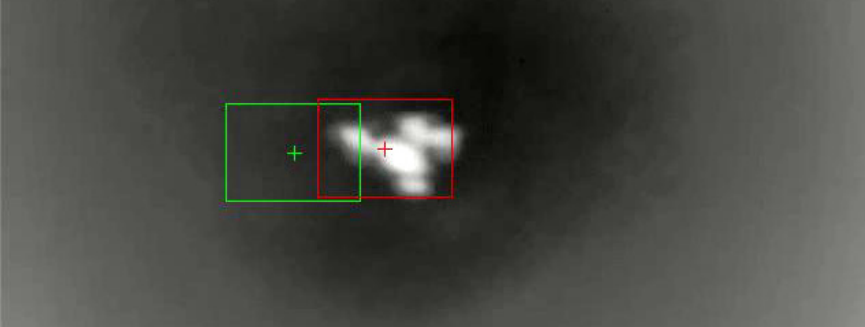}
   \includegraphics[width=1\linewidth]{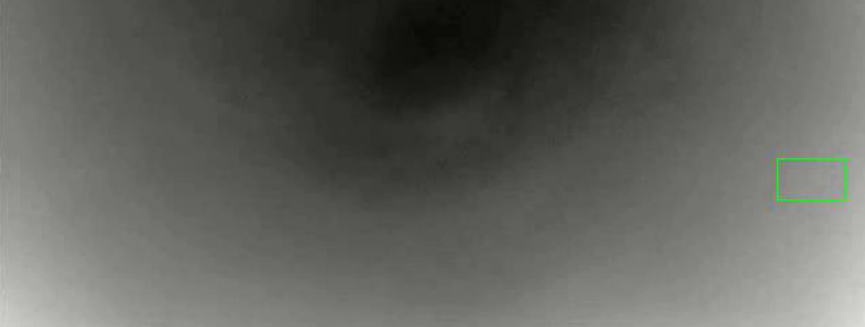}
   \caption{Some annotation errors in Anti-UAV dataset are showed in green bounding boxes and their corrected versions are showed in red bounding boxes. (a) Taken from 213th frame of IR\_20190925\_130434\_1\_4, meanwhile (b) is taken from 620th frame of IR\_20190925\_130434\_1\_9.}
\label{fig:annot_error}
\vspace{-1em}
\end{figure}

In order to assess the behavior of YOLOv3 on Anti-UAV Challenge dataset better, we have carefully inspected the outputs of the algorithm, especially the frames on which the algorithm fails, i.e. frames with low IoU, miss or false alarm. After this inspection, it can be easily noticed that there are significant amount of gross annotation errors, some of which are shown in Figure~\ref{fig:annot_error}. 

Since the dataset is composed of consecutive video frames, and only some of them have significant annotation errors, most of the time, it might be possible to recover those annotation errors by using temporal data and classical methods. Conventional template matching methods, such as cross correlation or phase correlation are quite effective with a high pointing accuracy for the short time periods, i.e. only a few frames. Even if the recent learning-based methods outperform such fundamental methods, in general, it should be reminded that template matching methods have high pointing accuracy performance as long as the pose changes and changes in background are not significant. As Anti-UAV Challenge dataset contains 30fps videos, the pose changes between consecutive frames can be ignored, and the changes in background could be eliminated manually. Moreover, even if the annotations are erroneous, as long as the annotation error is small with respect to the object size, those shifts do not affect template matching methods as the most of the template is still covered by the object of interest. 

In order to find the position of an object box (defined on frame $k$) at frame $k + 1$, we search the neighborhood of annotated object center on frame $k + 1$ with cross correlation. Let $u_{k+1}$ be displacement between annotated object center on frame $k + 1$ and the matching point of the template defined on frame $k$. This difference should have three components: annotation error on frame $k$, $w_k$; annotation error on frame $k + 1$, $w_{k+1}$; and the error of the matching algorithm $v_{k+1}$. For the first frame, there are two unknowns (annotation errors in x and y axes) and each new frame introduces four new unknowns (annotation and matching errors on x and y axes), resulting in a underdetermined linear system. During our initial attempts, we observe that minimum-norm solution of such an underdetermined system tends to assign most of the displacements between consecutive frames to matching errors. If the search range is large enough and pose change is not significant, the error of matching algorithm is usually small but it can cause some drift. In order to avoid the drift, we accumulate the displacements, fit a line to this cumulative displacement, and remove the resulting trend from the cumulative. 

During the experiments, we have observed that the search range is not large enough for some frames, but increasing search range might result in additional errors; therefore, we perform the annotation correction in two steps for the same search range (20 pixels). After this automatic correction, visual results of original annotations and automatically corrected annotations compared by a human operator, and better performing one is selected manually. Human operators preferred to use automatically corrected annotations for 66 videos over 100 thermal videos in dataset. For those 66 videos, the first and the second order statistics of difference between original annotations and corrected annotations on x- and y-axes are presented in Table~\ref{tab:realvscorrecteddiff}. When the corrected annotations are investigated, the annotation errors are mostly due to box shifts which are explained in Section 6. Therefore, the numerical values in Table~\ref{tab:realvscorrecteddiff} mainly correspond to parameters of shifted boxes. Corrected annotations and correction algorithm for thermal images of AntiUAV dataset are avaliable at \href{https://github.com/aybora/CVPR2020-Anti-UAV-OGAM-Correction/}{github.com/aybora/CVPR2020-Anti-UAV-OGAM-Correction/}

\begin{table}[h!]
\centering
\caption{Mean and standard deviations of difference and normalized difference with respect to width and height of bounding boxes between the center values of given and corrected annotations of 66 videos.}
\label{tab:realvscorrecteddiff}
\medskip
\begin{tabular}{|l|c|c|c|c|}
\hline
 & $\mu_x$ & $\sigma_x$ & $\mu_y$ & $\sigma_y$ \\
\hline
Diff. & 0.0970 & 2.729 & 0.0102 & 1.720 \\
Norm. Diff & 0.0022 & 0.0559 & 0.0015 & 0.0579 \\
\hline
\end{tabular}
\end{table}

\vspace{-1em}

\section{Experiments}

For data annotation, researchers generally either label the objects one-by-one for each image, or they make the labeling between some period of frames (e.g. labeling each 10th frame) and interpolate the bounding box values between the labeled frames by using a reliable tracker, especially for video annotation. Therefore, annotation error sources can be classified into two types: human-based and tracker-based faults. In the next part, both kinds of error sources are examined and their simulation results are presented. The performance of YOLOv3 with such simulated annotation errors is compared with error-free (original annotations) and corrected annotations mentioned in Section 4, and reported in the next part of the Section.

\subsection{Simulations of Various Annotation Errors}

\textbf{Additional boxes: }This type of error includes an extra box which does not contain any target. An additional box due to human fault should have a similar appearance with true objects and temporal consistency as a human tends to repeat the fault in consecutive frames. However, additional boxes due to tracker faults is due to either lack of object is visible/invisible decision mechanism, which generates random results without any temporal consistency or an erroneous decision of tracking algorithm which results in additional boxes having a similar appearance to true objects with temporal consistency. Therefore, in this study two types of additional boxes are generated: a) additional boxes at random positions without temporal consistency b) additional boxes initiated on one frame and tracked through consecutive frames to achieve temporal consistency.
  
  In order to insert $P$\% additional boxes without temporal consistency, $P$\% for the frames selected randomly and a box having a random position and random size is added. The position of the box is sampled from uniform distribution which covers the whole image, where as the size of the box is selected from a Gaussian distribution, whose mean and variance is set to mean and variance of object size in whole dataset. 
  
  Temporally consistent additional boxes should also have a similar appearance to true objects. To insert $P$\% temporally consistent additional boxes, for every 100 frames, we pick candidate additional boxes at random positions for the first (100 - $P$) frames. Then, for  simulating the visual similarity to true targets, candidate additional box with the highest variance is selected as the true objects have a different appearance from background which results in high variance. In order to simulate temporal consistency, selected additional box on the seed frame is tracked for $P$ frames with correlation tracker. 
  
  \textbf{Missing boxes:} A missing box error is simply due to the unavailability of the annotation of a true object. Completely random missing boxes are not expected, either due to human or tracker fault. Labeling people usually misses the objects due clutter or occlusion which is temporally consistent in general. Trackers have a similar behaviour, when they miss the target on one frame, they tend to miss the object in consecutive frames. To generate missing boxes with $P$\%, for every 100 frames, labeling of first (100 - $P$) frames is left as it is and the annotations are removed for the next $P$ frames to achieve temporal consistency. To examine whether this temporal consistency has a significant effect or not, temporally independent missing boxes are also simulated by selecting $P$\% of the frames independently for each video.
  
  \textbf{Shifted boxes: }A shifted box error is a slightly translated version of the true object box. As human eye cannot detect the object box very precisely in pixel or subpixel level, annotated boxes might be shifted by a few pixels. Trackers have a similar behaviour; even if they mark the true target, resulting bounding box might be shifted by a few pixels. Human errors can be assumed to have a zero mean Gaussian distribution. Tracker errors might be biased due to the drift behaviour of the tracker; however, in this work this effect is discarded. The shifted boxes are generated by adding zero mean Gaussian noise with the specified variance to original boxes without changing the size of the box. 
  
\begin{figure}[b]
\centering
   \includegraphics[width=1\linewidth]{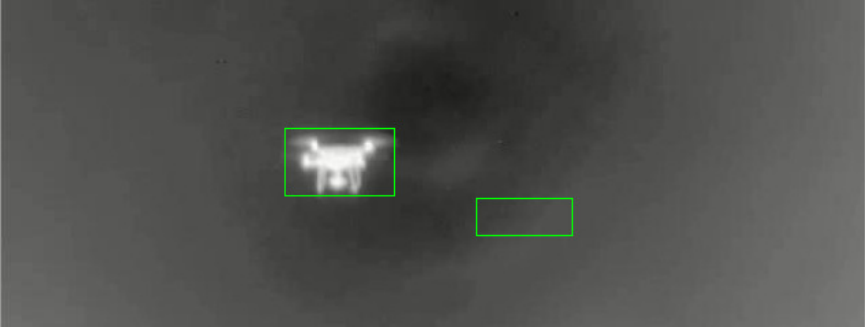}
   \includegraphics[width=1\linewidth]{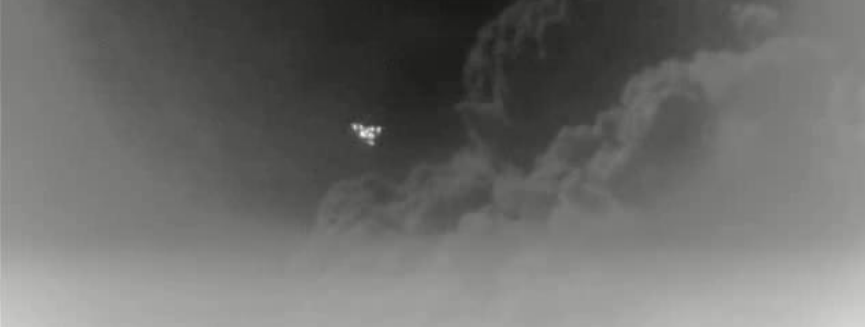}
   \includegraphics[width=1\linewidth]{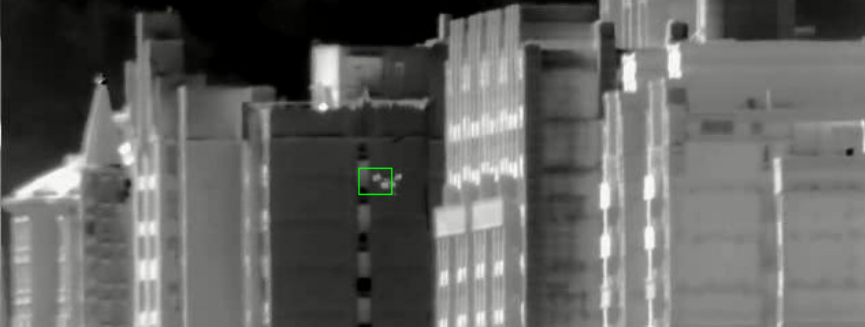}
   \caption{Visuals of simulated annotation errors: (a) additional box (b) missing boxes, (c) shifted box. }
\label{fig:errors}
\end{figure}

  Sample visuals for different types of annotation errors are presented in Figure~\ref{fig:errors}.

\subsection{Performance of YOLOv3 with Simulated Annotation Errors}

For all of the experiments presented in this section, the same training and validation sets are utilized. The simulated annotation errors are only applied to the training sets, and YOLOv3 is trained with erroneous annotations for each experiment independently. For the corrected annotation experiments, the network is trained with corrected annotations,  whereas the results are evaluated both with original and corrected annotations of validation set.

\textbf{Effect of additional boxes:} 
In the first experiment additional boxes with 25\% without temporal consistency are added to the training set. As shown in \textit{Additional Boxes (25\%)} column of Table~\ref{tab:compthrfixed}, when objectness threshold is fixed (0.5), hit rate is slightly increased with respect to training with original annotations, which slightly increases tracking accuracy and modified tracking accuracy as expected, since the number of misses decreases. However, the number of false alarms are increased from 2.4 FA/min to 9.7 FA/min. This result is probably due to a general increase trend in objectness scores. When objectness threshold is increased to fix the number of false alarms (2.4 FA/min for original annotations), hit rate is dropped by 3.4\%, tracking accuracy and modified tracking accuracy are dropped by 2.5\% with respect to the original annotations as shown in \textit{Additional Boxes (25\%)} column of Table~\ref{tab:compfafixed}. It can be concluded that adding completely random boxes of rate 25\% is not sufficient to create some pattern that causes the network to learn false positives, it rather forces the network to generate higher objectness scores, which is also supported by the selected objectness threshold of 0.72 to get the same number of false alarms.

When additional boxes with 50\% without temporal consistency are added to the training set, again the false alarm rate increases significantly as shown in \textit{Additional Boxes (50\%)} column of Table~\ref{tab:compthrfixed}. However, in this case, hit rate and tracking accuracy are decreased. Apart from forcing the network to increase the objectness scores, such a large number of additional boxes seems to detoriate the generalization capacity of the network. To fix the number of false alarms (2.4 FA/min for original annotations) objectness threshold should be increased to 0.68 as shown in Table~\ref{tab:compfafixed}. In this case hit rate is dropped by 5.3\%, tracking accuracy and modified tracking accuracy are dropped by 6.6\%.

The results for temporally consistent additional boxes of 25\% is shown in \textit{Tmp.Cons.Add.Box. (25\%)} columns of Tables~\ref{tab:compthrfixed} and ~\ref{tab:compfafixed}. When compared to \textit{Additional Boxes (25\%)} column of Table~\ref{tab:compfafixed}, the performance is better for temporally consistent additional boxes and objectness threshold to fix the number of false alarms is closer to original threshold. These results indicate that the network finds it easier to reject these consistent additional false alarms which is not expected. It can be concluded that the proposed temporally consistent additional box generation method does not work as expected and failed to generate generalizeable additional boxes.

\textbf{Effect of missing boxes:} When missing boxes of \%25 without temporal consistency is introduced as the annotation error, hit rate and tracking accuracy decrease as well as the number of false alarms as shown in \textit{Missing Boxes (25\%)} column of Table~\ref{tab:compthrfixed}. When objectness threshold is set to fix the number of false alarms, hit rate is decreased by only 0.3\% and tracking accuracy is decreased only by 0.5\% as shown in  \textit{Missing Boxes (25\%)} column of Table~\ref{tab:compfafixed}. For the missing boxes without any temporal consistency, the network is still able to generalize the appearance of the object; however, objectness scores tend to decrease.

When missing boxes with temporal consistency is introduced as the annotation error, the detection performance decreases significantly as shown in \textit{Tmp.Cons.Mss.Box. (25\%)} and \textit{Tmp.Cons.Mss.Box. (50\%)} columns of Tables~\ref{tab:compthrfixed} and ~\ref{tab:compfafixed}. Even for the same rate of missing boxes (25\%) performance is degraded significantly. When missing boxes without temporal consistency is applied, the only effect is introducing false negatives to the training set; however, when missing boxes have temporal consistency, apart from false negatives certain poses of the object are excluded from training set. It can be concluded that, if one has to make a decision between temporally consistent false positives and temporally consistent false negatives in training set; it is better to choose temporally consistent false positives.

\textbf{Effect of shifted boxes:} For shifted boxes two different alternatives are evaluated: standard deviation of Gaussian noise is set to a fixed value (1.5 pixels) to simulate tracker errors and a 10\% of object size to simulate human faults. As the average size of the objects in Anti-UAV Challenge dataset is 50 pixels in width, the second one corresponds to a standard deviation of 5 pixels. As shown in \textit{Shifted Boxes ($\sigma=1.5$)} and \textit{Shifted Boxes ($\sigma=10\%$)} columns of Tables~\ref{tab:compthrfixed} and ~\ref{tab:compfafixed}, shifted boxes decrease the performance significantly. Shifted boxes result in lower objectness scores in general. When the objectness threshold is set to generate 2.4 FA/min, for the noise of 1.5pixels standard deviation the detection outputs has 3.6\% lower pointing and hit rate is decreased by 5.1\%. It should be remembered that hit rate is a thresholded version of pointing accuracy, i.e. low pointing accuracy causes a decrease in IoU and detection result is recorded as a miss due to low IoU.

\textbf{Effect of combined errors:} Finally, 25\% temporally consistent missing boxes, 25\% additional boxes without temporal consistency and shifted bounding boxes with $\sigma=10\%$ cases are combined to simulate an extreme annotation error case. The results can be seen in \textit{Combined} columns of Tables~\ref{tab:compthrfixed} and ~\ref{tab:compfafixed}. As expected, the performance of YOLOv3 significantly degraded for such an extreme case.

\textbf{Effect of annotation correction:} 
Up to this point, it is assumed that the published annotations of Anti-UAV Challenge dataset is error-free; however, as stated in Section 5, there are significant annotation errors within the dataset. The proposed annotation correction method is applied to whole dataset and for 66 of 100 videos, corrected annotations are preferred by human operators. 

When the network is trained with the proposed corrected annotations and the results are evaluated with the original annotations, the performance is increased as shown \textit{Corrected Training} columns in Tables~\ref{tab:compthrfixed} and ~\ref{tab:compfafixed}. For the fixed objectness threshold, false alarm rate is slightly increased as well as hit rate and pointing accuracy. To make a fair comparison, objectness score is set to generate same number of false alarms with original annotations case. As in Shifted Boxes($\sigma = 1.5$) case, objectness threshold is obtained quite close to 0.5, however, this threshold update has no effect on the other metrics. Even evaluated with the original annotations, training with corrected annotations increase the hit rate and pointing accuracy. This result supports the argument that the corrected annotations are better. Therefore, as a final experiment, performance of the corrected training set is evaluated with the corrected validation set, whose results support the conclusion about annotation errors in dataset. As shown \textit{Corrected Training+Val} column of Table~\ref{tab:compfafixed}, when the corrected training set is evaluated with corrected validation set the highest performance is observed.

\begin{table}[h]
\centering
\caption{Performance comparison of YOLOv3 on thermal images in terms of False Alarms (FA / minute), Hit Rate (HR \%), Tracking Accuracy (TA \%) and Modified Tracking Accuracy (MA \%) when different noise types are applied with different probabilities and objectness threshold is fixed to 0.5}
\label{tab:compthrfixed}
\medskip
\begin{tabular}{|l|c|c|c|c|}
\hline
 & FA & HR & TA & MTA \\
\hline
Original Annotations & 2.4 & 97.5 & 73.6 & 73.5 \\
Corrected Training & 3.0 & 98.0 & 74.8 & 74.7 \\
Corrected Training+Val & 2.9 & 98.8 & 76.3 & 76.2 \\
Additional Boxes (25\%) & 9.7 & 97.8 & 74.9 & 74.3 \\
Additional Boxes (50\%) & 18.8 & 95.6 & 69.4 & 68.6 \\
Tmp.Cons.Add.Box.(25\%) & 5.6 & 96.5 & 72.7 & 72.5 \\
Missing Boxes (25\%) & 0.3 & 94.1 & 71.3 & 71.3 \\
Tmp.Cons.Mss.Box.(25\%) & 1.0 & 83.2 & 62.5 & 62.4 \\
Tmp.Cons.Mss.Box.(50\%) & 0.9 & 34.7 & 27.2 & 27.2 \\
Shifted Boxes ($\sigma=1.5$) & 2.2 & 90.8 & 68.8 & 68.8 \\
Shifted Boxes ($\sigma=10\%$) & 1.1 & 29.9 & 23.3 & 23.3 \\
Combined & 2.3 & 71.2 & 54.2 & 54.2 \\
\hline
\end{tabular}
\end{table}

\vspace{-1em}

\begin{table}[h]
\centering
\caption{Performance comparison of YOLOv3 on thermal images; Objectness Threshold (TH), Hit Rate (HR \%), Tracking Accuracy (TA \%) and Modified Tracking Accuracy (MTA \%) when different noise types are applied with different parameters for the False Alarm Rate of 2.4FA/minute}
\label{tab:compfafixed}
\medskip
\begin{tabular}{|l|c|c|c|c|}
\hline
 & Th & HR & TA & MTA \\
\hline
Original Annotations & 0.50 & 97.5 & 73.6 & 73.5 \\
Corrected Training & 0.55 & 98.0 & 74.8 & 74.7 \\
Corrected Training+Val & 0.55 & 98.8 & 76.3 & 76.2 \\
Additional Boxes (25\%) & 0.72 & 94.1 & 72.1 & 72.0 \\
Additional Boxes (50\%) & 0.68 & 92.2 & 67.0 & 66.9 \\
Tmp.Cons.Add.Box.(25\%) & 0.58 & 95.6 & 72.0 & 72.0 \\
Missing Boxes (25\%) & 0.40 & 97.2 & 73.2 & 73.1 \\
Tmp.Cons.Mss.Box.(25\%) & 0.38 & 90.8 & 67.9 & 67.8 \\
Tmp.Cons.Mss.Box.(50\%) & 0.30 & 56.0 & 43.2 & 43.1 \\
Shifted Boxes ($\sigma=1.5$) & 0.48 & 92.4 & 70.0 & 69.9 \\
Shifted Boxes ($\sigma=10\%$) & 0.30 & 87.8 & 64.8 & 64.7 \\
Combined & 0.49 & 72.8 & 55.4 & 55.3 \\
\hline
\end{tabular}
\end{table}

Finally, average IoU between corrected and original annotations are compared using Tracking Accuracy metric. TA is found 86.4\% in 66 corrected videos. It can be deduced that a perfect tracking algorithm which always gives correct results cannot have a tracking accuracy higher than 86.4\% on CVPR-2020 Anti-UAV Challenge test-dev dataset.

\section{Conclusion}
In this paper, the performance of a state-of-the-art object detector, YOLOv3, is evaluated for UAV detection problem which can also be used as a baseline for detection-based tracking methods. The YOLO Network is trained with Anti-UAV Challenge dataset to detect UAVs, and based on the results, it performs relatively well. While the detection performance is yielding relatively high hit rates and small false alarms; the tracking performance is not as good as the detection performance in terms of tracking accuracy or IoU. The tracking performance can be improved by utilizing the temporal information, even by employing some classical tracking techniques, such as a conventional Kalman filter that takes measurements from YOLO detector.

The performance of YOLOv3 is also tested on Anti-UAV Challenge dataset with different erroneous annotations, which is a typical problem in practice. The results are compared with a previous work~\cite{Chadwick} which is performed on KITTI dataset. Since small targets are already are quite difficult to detect, it is observed that the annotation errors degrades the performance much severely than that of KITTI dataset, especially for missing boxes. Moreover, the changes in objectness scores are quite noticeable when those annotation errors exist. Additional boxes increase the objectness score, while the missing boxes decrease it. Therefore, for a fair comparison, one of the metrics should be fixed and the other ones should be compared. 

There are some annotation errors in Anti-UAV Challenge dataset that are observed during the experiments. In order to correct such erroneous annotations, a correlation tracker is employed and the given annotations are updated in such a way that when an annotated object in one frame is searched in the next frame, the location with highest correlation score is the center of the annotated object of the next frame. Then, for each video, corrected annotations and the original ones are compared by human operators to select the annotation for that video. After such a correction mechanism, human operators preferred the corrected annotations for 66 videos out of 100 sequences.

Finally, it is observed that the corrected annotations increase both detection performance and tracking performance in terms of hit rate, false alarm rate and tracking accuracy. While the tracking accuracy is calculated 73.6\% in original annotations, it increases up to 74.8\%, in case of the corrected training set and original validation set being employed. Such a result reveals the success and necessity of the proposed annotation correction method. However, as the validation set also contains erroneous annotations and employed in the performance measurements, the increase in performance is limited. When the corrected training and validation sets are also employed, the tracking accuracy increases to 76.3\%. Therefore, to achieve fair results, the annotations of the challenge set should also be corrected.

\vspace{-1em}

\section{Acknowledgements}

\vspace{-1em}

This study is funded by ASELSAN Inc.

{\small
\bibliographystyle{ieee_fullname}
\bibliography{egbib}
}

\end{document}